\documentclass{article}
\usepackage{graphicx} 
\usepackage[margin=1in]{geometry}
\usepackage{booktabs}
\usepackage{amsmath}
\usepackage{amssymb}

\title{3D GANs and Latent Space: A comprehensive survey} 
\author{Satya Pratheek Tata, Subhankar Mishra \\ \\
        School of Computer Sciences,\\ National Institute of Science Education and Research (NISER)\\
        Bhubaneswar, Odisha}

\date{}

\begin{document}

\maketitle

\begin{abstract}
Generative Adversarial Networks (GANs) have emerged as a significant player in generative modeling by mapping lower-dimensional random noise to higher-dimensional spaces. These networks have been used to generate high-resolution images and 3D objects. The efficient modeling of 3D objects and human faces is crucial in the development process of 3D graphical environments such as games or simulations. 3D GANs are a new type of generative model used for 3D reconstruction, point cloud reconstruction, and 3D semantic scene completion. The choice of distribution for noise is critical as it represents the latent space. Understanding a GAN's latent space is essential for fine-tuning the generated samples, as demonstrated by the morphing of semantically meaningful parts of images. In this work, we explore the latent space and 3D GANs, examine several GAN variants and training methods to gain insights into improving 3D GAN training, and suggest potential future directions for further research.
\end{abstract}

\section{Introduction} \label{intro}
GANs generate realistic distributions from random noise. Starting from Ian Goodfellow's seminal paper \cite{goodfellow_generative_2014} in 2014, GANs have emerged into a hot research topic in the field of Machine Learning. Just in the year of 2021, 59,400 papers have been published on GANs and 2020 28,500 indicating its rage. Some of its areas of application include generating examples for image datasets \cite{sauer_stylegan-xl_2022, song_score-based_2020, vahdat_score-based_2021, dockhorn_score-based_2021,  kang_rebooting_2021, song_maximum_2021,  dhariwal_diffusion_2021, nichol_improved_2021, armandpour_partition-guided_2021, grcic_densely_2021}, human faces \cite{kim_soft_2022, ma_macow_2019,hazami_efficient-vdvae_2022, liu_pseudo_2021, menick_generating_2018}, cartoon and anime  characters\cite{sauer_projected_2021, karras_training_2020}, image-to-image translation \cite{hoyer_daformer_2022, chen_smoothing_2022, wang_cross-region_2021, zhang_prototypical_2021, zhou_context-aware_2021, schonfeld_you_2020, liu_learning_2019, katiyar_improving_2021, wang_high-resolution_2018}, text-to-image translation \cite{li_lightweight_2020, liu_fusedream_2021, zhang_stackgan_2018, li_manigan_2020, tao_df-gan_2022, zhu_dm-gan_2019, xia_tedigan_2021, xu_attngan_2018, wu_nuwa_2021} , super resolution \cite{ledig_photo-realistic_2017, bulat_super-fan_2018, chen_face_2017, guemes_super-resolution_2022}, photo inpainting \cite{banerjee_hallucinating_2020, han_face_2021, malesevic_photo-realistic_2019, hassanpour_e2f-gan_2022, olszewski_realistic_2017, yang_image_2018}, video prediction \cite{clark_adversarial_2019, luc_transformation-based_2021} and 3D object generation \cite{wu_learning_2016, li_3d_2019, ramasinghe_spectral-gans_2020, smith_improved_2017, nguyen-phuoc_blockgan_2020}.

The data generated by a GAN is realistic, but there are cases where fine-tuning is necessary. To this end, we understand the distribution associated with the random noise used to generate the data. We learn the disentangled representation of the latent space to find semantically meaningful directions and paths that help us morph semantically meaningful portions of images \cite{pajouheshgar_optimizing_2022, shen_interpreting_2020, shubham_learning_2021, yang_l2m-gan_2021}. Moving around in the latent space can help us embed images in the latent space\cite{yuan_embedgan_2021}, improve image realism \cite{wen_diamond_2021, dado_hyperrealistic_2022} and impose several regularizing constraints \cite{sendik_crossnet_2019}. 

Recent advancements in technology like augmented and virtual reality, self-driving vehicles, medical imaging etc, point in the direction of regularly using 3D objects on a large scale. Applying generative adversarial networks to 3D object generation helps us cut down the time and computational resources without depending on CAD models making 3D model generation and manipulation accessible to many. Building on the work on generating realistic images, 3D GANs aid in creating 3D models and realistic faces.

3D GANs map low dimensional random noise to 3D space and are used for 3D object generation. As mentioned earlier, 2D object generation using GANs captured a lot of attention but 3D GANs are still catching up in the task of 3D object reconstruction. 3D GANs are currently used in for point cloud reconstruction \cite{fei_comprehensive_2022, guo_deep_2020, xiao_unsupervised_2022}, 3D face generation \cite{mukhiddin_generative_2021}, 3D object reconstruction \cite{wu_learning_2016, fu_single_2021, gwak_weakly_2017, wang_shape_2017, wu_learning_2018, yang_dense_2019}, semantic scene completion \cite{chen_3d_2019, wang_adversarial_2018}, design automation, physics-based validation and simulation \cite{shu_3d_2019, vallecorsa_3d_2019}, cosmological GAN emulation \cite{tamosiunas_investigating_2021}, cell shape modelling \cite{wiesner_generative_2019}, generating efficient 3D anisotropic medical volumes \cite{granstedt_slabgan_2021} and in material science \cite{jangid_3dmaterialgan_2020}.

This paper presents a comprehensive survey of 3D GANs, with a particular focus on algorithms and applications. To our knowledge, this is the first comprehensive review paper discussing 3D GANs. We explore the latent representation of GANs and their importance in identifying interpretable directions in the generative process. We provide a thorough review of 3D-GAN concepts, variants, and state-of-the-art methods for efficiently speeding up the training process of 3D GANs. We also discuss challenges and trends for future research.

The paper is structured as follows: Section \ref{intro} provides an introduction to the current state of GANs, latent space, 3D GANs, and their challenges; Section \ref{rel_sur} summarizes related surveys; Section \ref{gan_theo} explores the GAN concept in subsection \ref{idea} and the latent space in subsection \ref{latent}; Section \ref{3d_gan} introduces 3D GANs, with subsection \ref{why_3dgan} exploring the need for 3D GANs, subsection \ref{3dg} discussing 3D-GAN and its improvements, subsection \ref{freq} introducing concepts frequently encountered when dealing with 3D modeling using 3D GANs, and subsection 
\ref{3dg_var} classifying 3D GANs based on the tasks they perform; Section \ref{train} discusses several 2D GAN variants and subsection \ref{clever} collects clever ways to train them; Section \ref{disc} discusses problems with both 2D in subsection \ref{disc2}, 3D GANs in subsection \ref{disc3}, and explores future directions in subsection \ref{disc4}; Section \ref{bib} includes a Bibliometric Analysis and finally Section \ref{la_fin} concludes the survey.
\section{Related surveys} \label{rel_sur}

In order to keep up with the recent advancements in GANs, several survey papers were published, but to the best of our knowledge, this is the first that talks about latent space exploration, manipulation and 3D GANs. 

To obtain a comprehensive understanding of 2D GANs; theory, algorithms and applications, refer table \ref{tab:2related_papers}. The surveys cover areas ranging from theory, challenges, algorithms, applications, recent advances \cite{wang_generative_2017, gui_review_2020, saxena_generative_2020, rizvi_spectrum_2021, pan_recent_2019, liu_generative_2020, hong_how_2020, hitawala_comparative_2018, creswell_generative_2018, aggarwal_generative_2021}, convergence \cite{barnett_convergence_2018}, stability \cite{jabbar_survey_2020, wiatrak_stabilizing_2020, zhang_self-attention_2019}, augmentation \cite{wang_survey_2020, shorten_survey_2019}, design improvements and optimizations \cite{pan_loss_2020, kurach_large-scale_2019}, GAN inversion \cite{xia_gan_2021}, face generation \cite{kammoun_generative_2022} and medical image analysis \cite{alamir_role_2022}. There are no surveys that exclusively focus on latent space exploration and manipulation to the best of our knowledge.

There are very few surveys on 3D GANs; on 3D face generation \cite{mukhiddin_generative_2021}, single image 3D reconstruction \cite{fu_single_2021} and predominantly deal with point clouds \cite{fei_comprehensive_2022, guo_deep_2020, xiao_unsupervised_2022}. Refer table \ref{tab:3related_papers} for related surveys on 3D GANs that cover point clouds, 3D face generation and 3D reconstruction from a single image.

Here is an improved version of the LaTeX code for better visibility:

\begin{table}
    \caption{Related surveys: 2D GANs}
    \label{tab:2related_papers}
    \begin{tabular}{p{0.4\textwidth} p{0.4\textwidth} p{0.1\textwidth} }
        \toprule
        \textbf{Category} & \textbf{Focus} & \textbf{Paper}\\
        \midrule
        Algorithms, theory, challenges and applications & Theory and challenges & \cite{wang_generative_2017}\\
         & Algorithms and applications & \cite{liu_generative_2020}\\
         & Recent developments and applications & \cite{rizvi_spectrum_2021}\\
         & Theory and applications & \cite{aggarwal_generative_2021} \\
         & GAN theory and applications & \cite{creswell_generative_2018} \\
         & GAN architecture evolution & \cite{hitawala_comparative_2018} \\
         & Recent advances & \cite{pan_recent_2019} \\
         & GAN theory & \cite{hong_how_2020} \\
         & Theory, algorithms and applications & \cite{gui_review_2020} \\
         & Design improvements and optimization & \cite{saxena_generative_2020} \\ 
        \hline 
        Stability & GANs, shallow NNs and Reinforcement learning & \cite{zhang_stable_2017} \\
         & Theory and stability & \cite{jabbar_survey_2020} \\
         & Stability, ways of training & \cite{wiatrak_stabilizing_2020} \\
        \hline
        Convergence & GAN convergence theory & \cite{barnett_convergence_2018} \\
        \hline
        Augmentation & GAN assisted face data augmentation & \cite{wang_survey_2020} \\
         & Image augmentation & \cite{shorten_survey_2019} \\
        \hline
        Design improvements and optimizations & Regularization and normalization & \cite{kurach_large-scale_2019}\\
         & Loss functions & \cite{pan_loss_2020} \\
        \hline
        Computer vision & Architecture and applications & \cite{wang_generative_2020} \\
         & Image synthesis & \cite{shamsolmoali_image_2020} \\
        \hline
        GAN inversion & GAN inversion & \cite{xia_gan_2021} \\
        \hline
         Face Generation & Face generation GANs & \cite{kammoun_generative_2022} \\
        \hline
        Medical Image analysis & Recent advances in medical field & \cite{alamir_role_2022} \\
        \bottomrule
    \end{tabular}
\end{table}

\begin{table}
    \centering
    \caption{Related surveys: 3D GANs}
    \label{tab:3related_papers}
    \begin{tabular}{p{0.3\textwidth} p{0.325\textwidth} p{0.05\textwidth} }
        \toprule
        \textbf{Category} & \textbf{Focus} & \textbf{Paper}\\
        \midrule
        Point cloud  & Point cloud processing & \cite{fei_comprehensive_2022}\\
         &  Point clouds & \cite{guo_deep_2020} \\
         &  Point clouds &\cite{xiao_unsupervised_2022}\\
        \hline
        3D Face Generation & 3D Face generation & \cite{mukhiddin_generative_2021} \\
        \hline
        Single image 3D reconstruction  & 3D reconstruction from single image & \cite{fu_single_2021}\\
\bottomrule
    \end{tabular}
\end{table}

\section{General Adversarial Networks} \label{gan_theo}
Generative modelling involves the automatic discovery and learning of regularities in the input data to generate new examples almost indistinguishable from the input data. GANs are clever generative models that frame the problem as a supervised learning problem with two sub-models; the generator (G) and the discriminator (D). G and D are trained together in a zero-sum game until the generator fools the discriminator half the time. This section will explain the core GAN idea and explore the latent space.

\subsection{GAN idea} \label{idea}
Machine learning models can be discriminative models or generative models. The generative models can be further classified into explicit density models and implicit density models. GANs are implicit density generative models that do not explicitly learn the input data's underlying distribution but produce results without any explicit hypothesis using the generated examples to modify itself \cite{bengio_deep_2014}. 

\begin{figure}[ht]
    \centering
    \includegraphics[width=10cm]{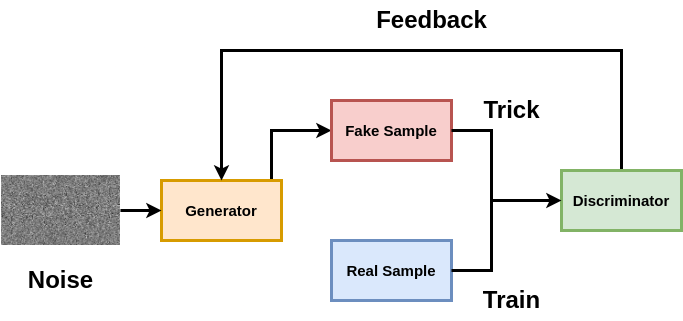}
    \caption{GAN architecture}
    \label{fig:gan}
\end{figure}

GANs use a generator and a discriminator to synthesize realistic images. A typical GAN contains two competing networks: the generator (G) and the discriminator (D) implemented using any differentiable system but many popular implementations use neural networks. The generator studies the input examples, $\emph{p}_{true}(x)$ and tries generating samples, $\emph{p}_{gen}(x)$ using an easy to model pre-defined prior distribution, $\emph{p}_{prior}(\mathbf{z})$ $(\mathbf{z}$ is noise) that resembles the input distribution and the discriminator is a classification model that distinguishes between true inputs from the fake ones. The architecture of a typical GAN is given in fig \ref{fig:gan}.

The GAN optimization is an example of a min max optimization.
\begin{equation}
        \min_G \max_D V(G,D) = \mathbb{E}_{x \sim \emph{p}_{true} (x)} [\log(D(x))] + \mathbb{E}_{\mathbf{z} \sim \emph{p}_{prior}(\mathbf{z})} [1 - D(G(\mathbf{z}))]
\end{equation}
where $\log D(x)$ is the cross-entropy between $[1\quad  0]^T$ and $[D(x)\quad  1-D (x)]^T$. Similarly, $\log (1-D(G(\mathbf{z})))$ is the cross-entropy between $[0\quad 1]^T$ and $[D(G(\mathbf{z}))\quad 1-D(G(\mathbf{z}))]^T$. 

For a fixed generator G, the optimal discriminator D is given by 
\begin{equation}
    D^*_G(x) = \frac{\emph{p}_{true}(x)}{\emph{p}_{true}(x) + \emph{p}_{prior}(x)}   
\end{equation}

Substituting $D^*_G(x)$ in (1) and after a bit of work, we get a cost function 
\begin{eqnarray*}
    C(G) &=& KL(\emph{p}_{true}\ ||\ \frac{1}{2}(\emph{p}_{true} + \emph{p}_{prior})) + KL(\emph{p}_{prior}\ ||\ \frac{1}{2}(\emph{p}_{true} + \emph{p}_{prior})) - 2\log2  \\
    &=& 2JS(\emph{p}_{true}\ ||\ \emph{p}_{prior}) - 2\log2
\end{eqnarray*}
, where $KL$ and $JS$ are Kullback-Leibler and Jensen-Shannon divergences respectively. 

\begin{figure}[ht]
    \centering
    \includegraphics[height=6cm]{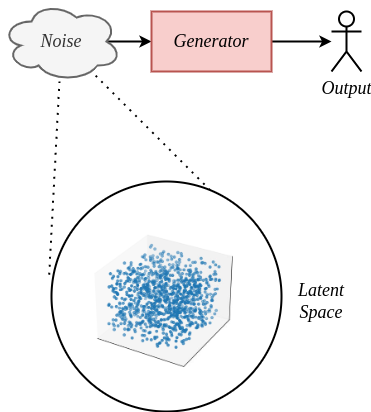}
    \caption{GAN's Latent Space}
    \label{fig:latent}
\end{figure}

The choice of distribution used to produce sampled noise vector as an input to train the generator is crucial as it represents the latent space, as seen in fig \ref{fig:latent}. Typically, the noise is from the normal distribution. The effect of the latent space dimension on GAN is in \cite{padala_effect_2020}. Work done on early GANs showed we could perform semantically meaningful vector space arithmetic and that certain directions in the latent space map to facial features \cite{radford_unsupervised_2016}. Building on this work \cite{shen_interpreting_2020} manipulate the latent space to edit images and open doors for further research.  

\begin{table}
    \caption{Questions and Answers about GANs}
    \label{tab:qna_gans}
    \begin{tabular}{p{0.4\textwidth} p{0.4\textwidth} p{0.1\textwidth} }
        \toprule
        \textbf{Question} & \textbf{Answer} & \textbf{Papers}\\
        \midrule
        Can GANs model any complex distributions? & No, cannot model/generate heavy-tailed distributions. & \cite{oriol_theoretical_2021}\\
        \hline
        Can GANs model discrete distributions? & Yes, using MaliGAN and improvements. & \cite{che_maximum-likelihood_2017, montahaei_dgsan_2021}\\
        \hline
        What can we learn about the GAN’s latent space? & Latent space is a Riemannian manifold & \cite{arvanitidis_latent_2021, chen_metrics_2018, kuhnel_latent_2018}\\
        & Disentangled representations & \cite{shen_interpreting_2020}\\
        & Semantically meaningful directions & \cite{voynov_unsupervised_2020}\\
        \hline
        What can we do by moving around in the latent space? & Easily compute difference between real and fake distributions & \cite{zhang_improving_2021}\\
        & Associate different regions of latent space to different layers of the generator & \cite{voynov_rpgan_2019}\\
        & Latent space Interpolation & \cite{wu_learning_2016}\\
        & Extend latent space interpolation & \cite{chen_homomorphic_2019}\\
        & Find interpretable paths & \cite{vahdat_score-based_2021}\\
        & Embed Images in Latent space & \cite{yuan_embedgan_2021}\\
        & Improve image realism & \cite{wiatrak_stabilizing_2020}\\
        & Impose regularizing constraints during unpaired image translation & \cite{shamsolmoali_image_2020}\\
        & Orthogonal feature variation & \cite{balakrishnan_rayleigh_2022}\\
        \hline
        Applications & Generate Hyper-realistic images & \cite{dado_hyperrealistic_2022}\\
        & Face editing & \cite{shen_interpreting_2020, yang_l2m-gan_2021, shubham_learning_2021}\\
        & Local image editing & \cite{pajouheshgar_optimizing_2022}\\
        & Photo-realistic multiple attribute transformation with identity preservation & \cite{zhuang_enjoy_2021}\\
        & Inspire creators and designers & \cite{roziere_inspirational_2021}\\
        & Open-ended primitive visual concept vocabulary & \cite{schwettmann_toward_2021}\\
        & Cosmological GAN emulation & \cite{tamosiunas_investigating_2021}\\
        & Expression label based image generation & \cite{wang_expression-latent-space-guided_2021}\\
        & Predict and compress videos in latent space & \cite{liu_deep_2021}\\
        & Procedural content generation of game levels & \cite{Kumaran_Mott_Lester_2020, fontaine_illuminating_2021}\\
        & Material science & \cite{jangid_3dmaterialgan_2020}\\
        & Multi-scale 3D point generative model & \cite{egiazarian_latent-space_2020}\\
        \bottomrule
    \end{tabular}
\end{table}
\subsection{Distributions and latent space} \label{latent}

With the ability to explore the latent space to edit images, researchers have begun to learn disentangled representations of the latent space, making it a significant research problem in current computer vision research. Now that we have an understanding of what a GAN is, we will shift our focus to the latent space and the nature of distributions modeled by a GAN by answering the following questions summarized in the table \ref{tab:qna_gans}:
\begin{itemize}
    \item \textbf{Can GANs model any complex distributions?}: In \cite{oriol_theoretical_2021}, the authors dispel the previously held notion of GANs being able to successfully model any probability distributions by showing that GANs cannot generate heavy-tailed distributions. 
    \item \textbf{Can GANs model discrete distributions?}: Despite GANs successfully capturing continuous distributions, they fail to capture discrete distributions. To solve this issue, the authors of \cite{che_maximum-likelihood_2017} introduce Maximum-Likelihood Augmented Discrete GANs (MaliGAN). Training the discriminator remains the same, but the authors introduce a novel objective for optimizing the generator using importance sampling. This brings the training procedure closer to maximum likelihood (MLE) training of auto-regressive models, increasing stability due to less variance in the gradients.
    \item \textbf{What kind of distributions can GANs model?}: In \cite{chen_statistical_2020}, the authors consider the data distributions in Holder space with minimal assumptions and provide some statistical guarantees regarding estimating data distributions using GANs. They view the inner maximization problem as an Integral Probability Metric (IPM); hence training a GAN boils down to minimizing the IPM between generated and input distributions. By doing so, we obtain the empirical estimator of input distribution as the pushforward distribution of easy-to-sample latent distribution under the optimal generator. 
    \item \textbf{What can we learn about the GAN's latent space?}: Any deterministic generative model can be modelled as a surface if its generator is sufficiently smooth. In the case of GANs, we model its latent space as a Riemannian manifold \cite{arvanitidis_latent_2021, chen_metrics_2018, kuhnel_latent_2018}. By performing linear transformations on the latent space, we can learn disentangled representations \cite{shen_interpreting_2020}. Using unsupervised learning, we can understand the semantically meaningful directions in the latent space \cite{voynov_unsupervised_2020} to appreciate the generative process better.
    \item \textbf{What can we do by moving around in the latent space?}: We can use unsupervised learning to easily and intuitively find interpretable paths in the latent space of a pre-trained GAN \cite{tzelepis_warpedganspace_2021} to control the generative process by learning non-linear warpings on the latent space, each one parametrized by a set of Radial Basis Function-based (RBF-based) latent space warping functions, and where each warping gives rise to a family of non-linear paths via the gradient of the function. Properly selecting paths can change specific attributes in an image, which helps generate intermediate images between the two domains. We can extend latent space interpolation during image translation by including their intermediate region \cite{chen_homomorphic_2019}. This is possible as many paths connecting two sample points exist in a flat, smooth latent space. By choosing Rayleigh Eigen Directions (REDs) in a generative model's latent space, we can generate appropriately curved paths that orthogonalize feature changes in an image \cite{balakrishnan_rayleigh_2022}. Associating different layers of the generator with different regions of the latent space can help us naturally interpret the latent space as seen in \cite{voynov_rpgan_2019} where the authors introduce Random Path GAN (RP-GAN), whose latent space consists of random paths in a generator network. By learning and leveraging an additional low-dimensional latent space which retains information from the original high-dimensional distributions, we can efficiently compute the difference between real and fake data distributions, as seen in \cite{zhang_improving_2021}. We first map the complex data distributions to simple Gaussian distribution and then find an appropriate divergence measure for the transformed real and fake data distributions. Directly embedding images in a GAN's latent space \cite{yuan_embedgan_2021} is possible by performing a latent walk starting from a randomly generated latent code in GAN to adjust the generation result. By finding a direction that best aligns with improved photo-realism in the latent space, we can improve image realism in a pre-existing low-complexity GANs \cite{wen_diamond_2021}. During unpaired image translation using GANs, we learn an image translation operator. We can impose several regularizing constraints on this learnt operator by using a pair of GANs with a pair of translators between their latent spaces to enable cross-consistency \cite{sendik_crossnet_2019}.
    \item \textbf{Some applications of exploring GAN's latent space}: Using functional magnetic resonance imaging (fMRI) data of the participant as they perceive face images created by the generator, we can generate hyper-realistic images using a decoder to predict regions of the GAN's latent space that correspond to hyper-realistic images \cite{dado_hyperrealistic_2022}. Using the InterFaceGAN that interprets the regions in the latent space corresponding to different semantics, we can semantically edit human face images \cite{shen_interpreting_2020}. Using the L2M-GAN, we can end-to-end edit of local and global attributes of a person's face \cite{yang_l2m-gan_2021}. We can learn a deep reinforcement policy on the latent space of a pre-trained GAN to manipulate a person's age without changing other semantically meaningful facial attributes \cite{shubham_learning_2021}. Using a pre-trained segmentation network and Locally Effective Latent Space Direction (LELSD) to optimize latent directions, we can evaluate image-edit locality to perform local editing  \cite{pajouheshgar_optimizing_2022}. This technique does not depend on the choice of dataset or GAN architecture in solving ambiguity between semantic attributes in a GAN-based localized image edit. We can learn more than one attribute transformation to integrate transformation function training with attribute regression increasing photo-realism, and identity preservation \cite{zhuang_enjoy_2021}. Using a latent space GAN and a Laplacian GAN, we can create a multi-scale 3D point cloud generation model that can generate 3D point clouds at increasing levels of detail by learning the Laplacian pyramid representation of the latent space \cite{egiazarian_latent-space_2020}. Designers and creators can derive inspiration and generate controlled outputs from their choice dataset by using a simple strategy involving many optimization steps to find optimal parameters in the model's latent space \cite{roziere_inspirational_2021}. We can associate words with primitive visual concepts and these concepts with a GAN's latent space to create open-ended vocabularies, which can be used as labels to generate images with expressions corresponding to the labels \cite{schwettmann_toward_2021}. Using Expression-Latent-Space-guided Generative Adversarial Network (ELS-GAN), we can control changes in expressions in an image \cite{wang_expression-latent-space-guided_2021}. We can compress individual frames in a video by training an autoencoder with a GAN. Using a ConvLSTM network, we can predict the next frame's latent vector. Combining the two, we can compress video sequences in a GAN's latent space \cite{liu_deep_2021}. Procedural Content Generation of levels in multiple distinct video games is possible by starting from a common gameplay action sequence and probing the common latent space \cite{Kumaran_Mott_Lester_2020}. Given a specific set of gameplay measures, we can extract several corresponding levels \cite{fontaine_illuminating_2021}. Given a 3D polycrystalline microstructure, we can use 3DMaterialGAN to recognize and synthesize morphologies that conform with the given structure up to individual grains without the need for 2D rendered images \cite{jangid_3dmaterialgan_2020}. Cosmological GAN emulators can be investigated by interpolating the latent space \cite{tamosiunas_investigating_2021}.

The works mentioned above suggest that a wealth of opportunities lie in the local analysis of the geometry and semantics of latent spaces leading to several applications.
\end{itemize}

\section{3D GANs} \label{3d_gan}
3D shape generation is a complex problem due to the higher dimension of the generator's output space. Previous attempts used CAD models, rendering engines, etc., making 3D modelling computationally expensive. A 3D GAN's generator maps a low-dimensional latent space to the space of 3D objects so we can sample objects without needing a reference image or a CAD model to explore its 3D object manifold making 3D modelling accessible.

In the subsequent subsections, we find that 2D GANs cannot directly help us generate 3D objects. We look at the first 3D GAN, \emph{i.e.} 3D-GAN and its improvements, then we familiarize ourselves with some of the frequent concepts in 3D modelling, and we will end this section by looking at various 3D GAN variants.

\subsection{Prelude} \label{why_3dgan}
Before we introduce 3D GANs, we want to know if 2D GANs can give us some insight into the 3D structure of their input images. In \cite{pan_2d_2021}, the authors answer the question, "Is it possible to reconstruct the 3D shape of a single 2D image by exploiting the 3D-alike image manipulation effects produced by GANs?" with a yes using their unsupervised approach, GAN2Shape, by showing that when existing 2D GANs are trained only on images, they could accurately reconstruct its 3D shape for objects belonging to several categories including human faces, cars, buildings among others without the need for a 2D keypoint or 3D annotations. They also found that GAN2Shape works with images from the wild using an improved GAN inversion strategy.

GAN2Shape fails for complex objects like horses where the viewpoint and lighting variations cannot be captured using simple convex priors resulting in the inaccurate inference of 3D shapes. GAN2Shape's 3D mesh uses a depth map which cannot model off-camera portions.

This is why we need 3D GANs which work with sophisticated shapes and model the off-camera portions of images.

\subsection{3D-GAN and its improvements} \label{3dg}

\subsubsection{\textbf{3D-GAN}}
\begin{figure}[ht]
    \centering
    \includegraphics[width=10cm]{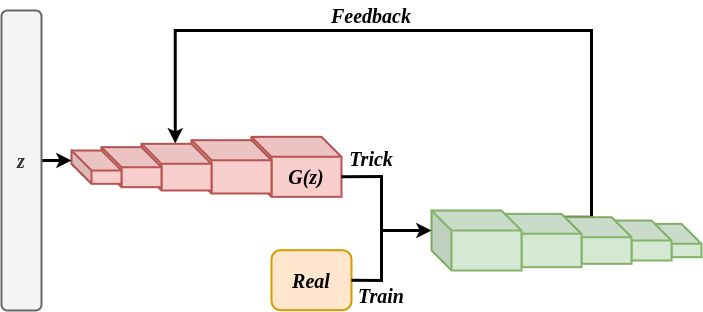}
    \caption{3D-GAN's architecture}
    \label{fig:3D GAN}
\end{figure}

A 3D GAN is conceptually similar to the vanilla GAN and can generate realistic and varied 3D shapes. It was introduced in \cite{wu_learning_2016}, where the authors introduce 3D-GAN and its extension 3D-VAE-GAN by using 3D convolutional networks to represent discrete volumetric structures and GANs. 3D-GAN's generator implicitly captures the object structure to generate high-quality 3D objects without a reference image or CAD models to explore the 3D object manifold. 3D-GAN's discriminator can be used as a robust unsupervised 3D shape descriptor and is helpful in 3D object recognition. Fig \ref{fig:3D GAN} shows 3D-GAN's architecture.
\subsubsection{\textbf{Improvements}}

Given a latent vector, a 3D-GAN can just map its latent space trained on examples having voxel grids. By designing its projection operator from a user-provided 3D voxel grid to a feature vector in the latent space of a 3D-GAN, we can overcome this limitation \cite{liu_interactive_2017}. This improvement ensures that the generated output is realistic while having a similar shape as the input but projected to that part of the manifold corresponding to realism.
 
3D Improved Wasserstein Generative Adversarial Network (3D-IWGAN) \cite{smith_improved_2017} can understand the detailed 3D shape of objects. It uses Wasserstein distance normalized with gradient penalization as a training objective, improving object generation joint shape distribution. It can be used to reconstruct 3D shapes from 2D images and for shape completion of occluded 2.5D range scans.

3D shape generation that sequentially transitions from coarse to fine-scale can be achieved by first using a primitive GAN, which uses primitive parts of a shape as attributes, followed by a 3D-VAE-GAN to add fine-scale details to the shape \cite{khan_unsupervised_2019}.
 
\subsection{Frequent concepts in 3D modelling using 3D GANs} \label{freq}
Given below are a few key concepts we frequently encounter while 3D modelling using GANs:
\begin{itemize}
    \item \textbf{Domain Translation}: Let A, B $\in \mathbb{R}^{w \times h \times 3}$ be two spatial domains encoding information as RGB values. Then the maps 
    \begin{align*}
        G_{x\rightarrow y}: x \rightarrow y, x,y \in A, B \\
        G_{y\rightarrow x}: y \rightarrow x
    \end{align*}
    translate between images having fixed width and height in domains A and B, respectively. 
    \item \textbf{Point clouds}: Point clouds are a set of data points or coordinates in three dimensions. Points in a point cloud are not ordered like pixels in a grid. Point sets are invariant under translation or rotation. The interaction of points, among others, influences the overall category of a point-set. The final shape having meaning is obtained by summing the points globally.

    \item \textbf{Voxelization}: A voxel is a 3D counterpart of the 2D pixel. The voxel grid is a 3D geometry type defined on a regular 3D grid. Voxelization involves transforming a point nucleus into a voxel grid and estimating the geometries and attributes created or embedded by points inside voxels. As voxel grids are defined on a regular 3d grid, they have structured volumetric representation where we can perform 3D convolutions, which helps us convert point clouds into a voxel grid using models like \emph{VoxNet} \cite{stornaiuolo_3d_2020}. However, converting a point cloud representation to a voxel grid takes time.

    \item \textbf{Meshes}: A polygonal mesh is a set of vertices and polygonal elements that collectively define a three-dimensional geometrical shape approximating 3D surfaces. Points in a point cloud are discrete samples from some continuous surface. We can use meshes to approximate this surface by using the points are vertices to fit polygons.

    \item \textbf{Depth maps}: Depth maps or RGB-D images are the conventional RGB images with an additional depth channel. The depth represents the distance between the image plane and the corresponding object. We can apply convolutions on RGB-D images as they are ordered structures. If we know the intrinsic properties of the camera used, we can recover the corresponding 3D point cloud representation. 
 
\end{itemize}

There are several 3D GAN architectures that translate one domain to another. For example, \cite{li_point_2018} is an image to point cloud GAN, \cite{leal-taixe_image--voxel_2019} is an image to voxel GAN, \cite{pemasiri_im2mesh_2021} is an image to mesh GAN, \cite{milz_points2pix_2019} is a point cloud to image GAN.

\subsection{3D GANs variants} \label{3dg_var}
The original 3D GAN formulation cannot be directly used to solve challenging problems in 3D modelling and other fields. There have been several advancements in how we use 3D GANs, either
modifications to the generator network and/or the discriminator network, its formulation or tweaks
to the loss functions to speed up the GAN training process and/or to obtain stable results. A few named 3D GANs are listed in table \ref{tab:3dvar}. We will classify several 3D GANs based on the task they perform here:

\begin{itemize}
    \item \textbf{Understanding and generating point clouds}: tree-GAN's generator is a tree-structured convolutional network (TreeGCN), and its discriminator uses Frechet point cloud distance. With this, a tree-GAN can use information from parent nodes to boost its representation features to produce accurate 3D point clouds \cite{shu_3d_2019}. Using a combination of encoder-decoder for image segmentation, we can perform multi-modal domain translation between 3d point clouds and images \cite{milz_points2pix_2019}. In \cite{achlioptas_learning_2018}, the authors introduce two models, Raw point cloud GAN (r-GAN) for learning the representations of raw point clouds and Latent-space GAN (l-GAN), where the raw point cloud input is passed through a pre-trained autoencoder first. Both these models aim to provide a learning representation of 3D point clouds. Using a Periodic Implicit GAN ($\pi$-GAN), we can learn 3D representations without supervision from images \cite{chan_pi-gan_2021}. We do this by conditioning input noise on an implicit radiance field represented by a SIREN network \cite{sitzmann_implicit_2020}. This conditioned radiance field takes 3D locations and 2D viewing directions as input to produce radiance and volume density. Radiance is view-dependent, while the volume density is not. Layers in the SIREN network are conditioned using feature-wise linear modulation (FiLM). To improve training speed and deal with complex computations in a 3D GAN, a progressive, growing strategy \cite{karras_progressive_2018} is introduced.
    \item \textbf{Single image-based 3D reconstruction}: Coming with the complete 3D reconstruction that fills missing or occluded regions from a single arbitrary voxel representation of depth view image of an object is possible using 3D-RecGANs \cite{yang_3d_2017}. The authors combine auto-encoders with conditional GANs to accurately infer fine-grained 3D structures in the voxel space. Using the correspondences between 2D silhouettes of multiple objects in a single colour image and frustum slices in a camera, we can predict a voxel model with a Z-GAN \cite{leal-taixe_image--voxel_2019}. Z-GAN's generator has skip connections between 2D and 3D feature maps to exploit pyramid-shaped voxels. This paper also introduces VoxelHome and VoxelCity datasets with 36,416 colour images, ground-truth voxel models, depth maps and camera orientations in 7 indoor and 21 outdoor settings. 
    \item \textbf{Incomplete or corrupted 3D reconstruction}: 3D reconstruction supervised using foreground masks is possible \cite{gwak_weakly_2017} by using a ray trace pooling layer to enable backpropagation and perspective projection to constrain reconstructed samples to the manifold of unlabeled real 3D shapes. A low-resolution 3D object can be completed using a 3D-ED-GAN combining a 3D-ED-GAN, and by combining it with a Long term RCN, we can produce high-resolution completions \cite{wang_shape_2017}.
    \item \textbf{2D projection consistent 3D shape generator}: To generate 3D shapes of objects without direct access to shapes consistent with several 2D projections, the generated results are projected to obtain several viewpoints, which are then fed to a prediction network trained on silhouette images with viewpoints to estimate viewpoints of the generated samples in a multi-projection GAN (MP-GAN) \cite{li_synthesizing_2019}. The space of possible viewpoints is discretized into 16 view-bins for robustness, and view prediction is treated as a classification problem to a vector of view probabilities. 
    \item \textbf{3D geometry aware image synthesis}: To generate high resolution 3D-geometry-aware images effectively from unposed images, we use GRAF \cite{schwarz_graf_2020}. It samples images at various scales using a patch-based discriminator to produce radiance fields. Another strategy is to use an effective and efficient tri-plane-based 3D GAN framework \cite{chan_efficient_2021}. It uses dual discrimination to promote consistency from several viewpoints while the generator is conditioned on poses to faithfully model attribute distributions dependent on the pose in the real-world datasets.
    \item \textbf{Generating texture models or 3DMMs}: In \cite{gecer_ganfit_2019}, the authors introduce a novel 3D Morphable Model (3DMM) in the form of a GAN Texture Model to provide excellent facial shape and texture reconstructions in arbitrary recording conditions from 2D images. They also show the results to be both photorealistic and identity preserving in both qualitative and quantitative experiments. Using MeshGAN, the first intrinsic GAN architecture operating directly on 3D meshes, we can generate high-fidelity 3D faces with rich identities, and expressions \cite{cheng_meshgan_2019}. 
    \item \textbf{Feedback learning}: By considering the generator network as an encoder and decoder, the spatial output from multiple discriminators can be used to provide feedback to the generator so it can improve on its previous generations using Adaptive Spatial Transform \cite{huh_feedback_2019}.
    \item \textbf{Meshes}: In industrial design, gaming, computer graphics and other digital art, automatically generating shapes based on meshes is necessary. Most of the current research is involved with voxel and point cloud generation, alienating itself from the design and graphics communities. MeshGAN, as mentioned above, is the first intrinsic GAN architecture operating directly on 3D meshes and can generate high-fidelity 3D faces with rich identities and expressions. To automatically generate shapes based on meshes, we use the signed distance function representation to generate detail-preserving three-dimensional surface meshes \cite{jiang_hierarchical_2017}. We can add colour channels into the output without supervision by first using a target 3D textured model data set to adapt the channels in the voxel inputs and then using marching cubes to translate the voxel-based models into a naive coloured mesh \cite{spick_naive_2020}. 
 
\end{itemize}

Generating 3D representations from low dimensional latent space is time-consuming, resource-hungry and prone to instability during the training process. We do not have many dedicated 3D GAN variants that aim to improve training speed and decrease instability, so in the next section, we look at 2D GAN variants and clever ways of training them to see if the same techniques can be used.

\begin{table}
    \caption{Named 3D GAN variants}
    \label{tab:3dvar}
    \begin{tabular}{p{0.1\textwidth}|p{0.2\textwidth}|p{0.25\textwidth}|p{0.3\textwidth}}
        \toprule
        \textbf{S.No.} & \textbf{Name} & \textbf{Key feature} & \textbf{Resulting improvement}\\
        \midrule
        1 & 3D-GAN \cite{wu_learning_2016} & Volumetric CNN & 3D point cloud generation\\
        \hline
        2 & 3D-VAE-GAN \cite{wu_learning_2016} & Additional encoder & 2D image to 3D point cloud generation\\
        \hline
        3 & 3D-IWGAN \cite{smith_improved_2017} & Wasserstein distance normalized with gradient penalization & Improved generation, can reconstruct 3D shape from 2D image and can complete occluded 2.5D range scans.\\
        \hline
        4 & tree-GAN \cite{shu_3d_2019} & Tree structured graph convolution generator & Use ancestor information to boost feature representation power.\\
        \hline
        5 & r-GAN \cite{achlioptas_learning_2018} & Directly operates on point clouds & Learns raw point cloud representations\\
        \hline
        6 & l-GAN \cite{achlioptas_learning_2018} & Pre-trained AE & Learns raw point cloud representations\\
        \hline
        7 & $\pi-GAN$ \cite{chan_pi-gan_2021} & SIREN\cite{sitzmann_implicit_2020} network, Feature-wise linear modulation and progressive growing & Unsupervised 3D representation learning and accelerated training\\
        \hline
        8 & 3D-RecGAN \cite{yang_3d_2017} & Combines auto-encoders with conditional GANs & Infers accurate and fine grained 3D structures\\
        \hline
        9 & Z-GAN \cite{gwak_weakly_2017} & Manifold constrained resonstruction & 3D reconstruction from masks\\
        \hline
        10 & 3D-ED-GAN \cite{wang_shape_2017} & Combines encoder-decoder GAN and long term recurrent convolutional network & Produce complete high quality of low resolution corrupted 3D object\\
        \hline
        11 & MP-GAN \cite{li_3d_2019} & Pre-trained view prediction network & Generated 3D shapes consistent with multiple 2D projections\\
        \hline
        12 & MeshGAN \cite{cheng_meshgan_2019} & Chebyshev convolutional filters & Directly operates on 3D Meshes\\
        \bottomrule
    \end{tabular}
\end{table}

\section{GAN variants and ways to train them} \label{train}

The original GAN formulation cannot be directly used to solve challenging problems in computer vision and other fields. There have been several advancements in how we use GANs, either modifications to the generator network and/or the discriminator network, its formulation or tweaks to the loss functions to speed up the GAN training process and/or to obtain stable results. A few of the significant variants are listed in the table \ref{tab:var}.

\begin{table}
    \caption{Major GAN variants.}
    \label{tab:var}
    \begin{tabular}{p{0.15\textwidth} p{0.15\textwidth} p{0.3\textwidth} p{0.3\textwidth}}
        \toprule
        \textbf{Group.} & \textbf{GAN Variants} & \textbf{Key feature} & \textbf{Resulting improvement}\\
        \midrule
        Improve quality  & LapGAN \cite{denton_deep_2015}& cGAN + Laplacian pyramid  & Better photo-realistic images\\ 
         & EBGAN \cite{zhao_energy-based_2017} & Auto-Encoder + GAN & Better convergence pattern and scalability to generate high-res image\\ 
         & BEGAN \cite{berthelot_began_2017}&Energy based, MCMC sampler & Improve distribution modeling and sample quality\\
         & Bayesian GAN \cite{saatchi_bayesian_2017}& Stochastic Gradient HMC & Capture diverse, complementary and interpretable representations \\
        \hline
        Generalization & f-GAN \cite{nowozin_f-gan_2016}& f-divergence loss& Generalize vanilla GAN\\ 
         & cGAN \cite{mirza_conditional_2014} & Conditioning G and D& Extends vanilla GAN, Multi-modal mapping\\
         & InfoGAN \cite{chen_infogan_2016}& Mutual information& Generalize cGAN, unsupervised disentangled representation learning\\  
         & BC-GAN \cite{abbasnejad_bayesian_2017}& Random generator function & Extends traditional GANs to Bayesian framework, naturally handles unsupervised, supervised and semi-supervised learning problems\\
        \hline
        Vanishing Gradients  & LSGAN \cite{mao_least_2017}& Least-square loss& Overcomes vanishing gradients\\
         & WGAN \cite{arjovsky_wasserstein_2017}& Wasserstein loss & Removes vanishing gradients, partially removes mode collapse\\ 
         & Improved WGAN \cite{gulrajani_improved_2017}& Gradient penalty on D & Strong modeling performance, stability\\ 
        \hline
        Stability  & McGAN \cite{mroueh_mcgan_2017}& Mean and covariance& Stability feature matching IPM\\
         & Cramer GAN \cite{bellemare_cramer_2017} & Cramer loss & Increased diversity, stability\\ 
         & Fisher GAN \cite{mroueh_fisher_2017}& Scale invariant IPM & Stable and time efficient training\\ 
         & MAGAN \cite{wang_magan_2017}& Adaptive hinge loss& Stability and performance\\ 
         & SAGAN \cite{zhang_self-attention_2019}& Self-attention, Spectral normalization & Effective in modeling long-range stability \\
         & RGAN \cite{sarraf_rgan_2021}& Rényi loss & Stability\\
        \hline
        Training speed & FastGAN\cite{zhong_improving_2020} & Follow the Ridge (FR) algorithm & Reduce generation time, improve generation quality\\
        \bottomrule
    \end{tabular}
\end{table}

\subsection{Clever tricks to train GANs} \label{clever}
We can select only the top k samples by simply zeroing out gradient contributions from the batch elements, which the discriminator deems least important to improve training time \cite{sinha_top-k_2020}.

The Lottery ticket hypothesis\cite{frankle_lottery_2019} showed that a large neural network could be pruned to find a sparse sub-network (the lottery ticket) that performs comparatively with a fraction of parameters when compared to the original network. In \cite{chen_data-efficient_2021}, the authors first exploit this neural network feature to find the lottery ticket in GANs and then perform the standard GAN training procedure. 

We find an interesting perspective in \cite{qin_training_2020}, where the authors claim instability in GAN training is due to the discretization of continuous dynamics in the gradient descent, and we need accurate integration to impart stability. They present a novel and practical view that frames GAN training as solving ODEs in the form of ODE-GAN, which uses numerical solvers to solve ODEs. 

In \cite{grassucci_quaternion_2021}, the authors introduce QGANs which exploit Hamilton's product by considering process channels as a single entity and easily capturing latent relations resulting in marked improvements to generation ability and also saving three-fourths of the model parameters. This improves performance and replication and saves computational resources making the GANs more accessible.

Instead of building and training GANs from scratch, the authors of \cite{sauer_projected_2021} make use of the pre-trained feature networks and modifications to the discriminator architecture to drastically decrease GAN training time, improve training stability and faithfully extract more information from the input dataset. This is done by projecting different layers of the pre-trained feature network using random $1\times1$ convolutions, then successively mapping the projected layers using random $3\times3$ convolutions and finally mapping each projection to its respective discriminator. 

Training a GAN using just a few images on a relatively inferior GPU opens a previously unexplored avenue for GAN applications. This can be done by adding a skip-layer channel-wise excitation module and a self-supervised discriminator trained to encode features \cite{liu_towards_2021} or by using a  discriminator with two branches for content and layout designed to judge the realism of scene layouts separately from their internal content \cite{sushko_one-shot_2021} or by using a set of experimentally verified best practices \cite{hinz_improved_2021}. 

Only using strong supervision vastly improves the upper bound of image generation quality at the cost of earlier model collapse. By combining strong supervision with moderate regularization, Omni-GANs prevent early model collapse in image-to-Image translation, colourization, and prior enhanced super-resolution \cite{zhou_omni-gan_2021}.

\section{Discussion} \label{discussion} \label{disc}
In this survey, we have looked at the theory of 2D and 3D Generative Adversarial Networks, explored what kinds of distributions a GAN, 2D or 3D can model, exploring the latent space and how to manipulate it, looked at the new and less discussed 2D GAN variants, justified why we need 3D GANs and classified 3D GANs based on the task they perform.

As with any successful model, GANs have several problems that need further research. We will look at them separately for 2D and 3D GANs, although there is an overlap between the two. 
\subsection{2D GANs} \label{disc2}
These issues are taken from several other surveys such as \cite{saxena_generative_2021} and \cite{jabbar_survey_2020}.
\begin{itemize}
    \item \textbf{Training instability}: In a GAN, G and D engage in an adversarial relation to reach Nash equilibrium where they are simultaneously \cite{goodfellow_generative_2014}. Instability arises when there are no coordinated updates to their cost functions.
    \item \textbf{Mode collapse}: It happens when G always produces an identical output; either generated output lacks diversity or generates only a subset of the input distribution. 
    \item \textbf{Vanishing gradients}: When the gradients of the generator with respect to its earlier layers start becoming small, the initial layers in the network stop learning. D rejects generated samples with confidence because G does not share any information with D, which harms its learning capacity.
    \item \textbf{Lack of universal evaluation metrics}: Despite the GAN's success, it is still challenging to rank evaluation metrics \cite{borji_pros_2018} as the term evaluation lacks a consensus definition.
    \item \textbf{Internal covariate shift}: Arises due to a change in input distribution forcing the layers in the middle to adapt and counter the shift, slowing training.
    
\end{itemize}

\subsection{3D GANs} \label{disc3}
These issues relate to Single image-based 3D object reconstruction from \cite{fu_single_2021}.
\begin{itemize}
    \item \textbf{Shape complexity of objects}: Different objects belonging to different classes have different properties across scales in the reconstruction process. This means that training on the same class of objects produces different results when trained jointly with other object classes as the model is busy learning additional inter-layer connections while ensuring it generates unique results when trained on the same object class.
    \item \textbf{Uncertainty of objects}: Given an image, coming up with its 3D reconstruction is difficult due to three reasons; absence of prior knowledge, information loss, and non-unique reconstruction assumptions. 
    \item \textbf{Memory requirements and calculation time}: The model parameters could build in size, delaying fine-grained reconstruction.
    \item \textbf{Lack of training datasets}: While reconstructing, the current models improve their ability to recognize patterns \cite{tatarchenko_what_2019} instead of improving their ability to reconstruct as the test and train could be very similar, which is seldom the case with in-the-wild datasets.
\end{itemize}

\subsection{Future work} \label{disc4}

2D GANs have made much progress in the task of 2D object reconstruction. Despite that, they still suffer from unstable training, vanishing gradients, mode collapse, internal covariate shift and Lack of universal evaluation metrics mentioned in the previous section. Future research must focus on minimizing the effect of the problems mentioned above, if not eliminating them. Whereas 3D GANs are nowhere near the current state-of-the-art models for the task of 3D object reconstruction, 3D object detection, 3D human pose estimation and their subtasks. The researchers could choose to follow the footsteps of 2D GANs to improve 3D GANs or chart their way. Considering the leaps by 2D GANs and the recent strides by 3D GANs, we have identified some broad areas for future work. 

\begin{itemize}

\item \textbf{Lightweight 3D GAN architectures}: Several papers proposed several tweaks to either promote high-resolution generation or highly diverse generation but not both of them together while not suffering from vanishing gradients, mode collapse or introduction of unwanted artifacts. Future researchers could focus on this problem. Training a GAN or a 3D GAN takes time and requires expensive equipment, hence centralization. Tweaking the architecture to decrease training time on less expensive equipment helps democratize and decentralize its benefits to a more significant number of people. Researchers could focus on this issue.
 
\item \textbf{Agnostic 3D GAN models}: A 3D GAN's generator generates output belonging to a particular 3D representation. Future research could help us generate 3D representation agnostic models to generate the final output in all different 3D representations. Despite having a few papers that deal with generating high-quality 3D models at all scales, 3D GANs fail to generate high-quality 3D models at all scales for several real datasets. Further research could help solve this issue.

\item \textbf{Domain knowledge infused 3D GAN models}: Using GANs, including 3D GANs, alone, we have produced some beautiful results, but researchers could combine them with knowledge acquired from other fields to explore and solve previously unexplored problems. Unlike the previous expectation, some applications such as self-driving \cite{wang2022sfgan} could gain a lot from domain-specific generated adversarial samples.

\item \textbf{Adversial 3D GANs}: GANs are used by cybersecurity professionals to generate robust and scalable adversarial samples under the supervision of system designers as GANs fail to consider all the potential adversarial samples for the given system by themselves \cite{huang_context_2018, dziugaite_training_2015}. Not just that, the rapid changes to the systems make high-quality sample generation a difficult task. Researchers could aim to solve this issue to uphold the integrity of our networks.

\item \textbf{Ethical 3D GANs} The recent advances in GAN-generated images, videos and 3D models made image morphing, deepfakes and realistic human face generation even more accessible. This has several ethical, social, political and economic ramifications. Researchers could work on ensuring the net damage done on the ethical, social, political and economic fronts is minimized while advancing the generative power of GANs.  
\end{itemize}

\section{Bibliometric Analysis} \label{bib}
\begin{figure}[ht]
    \centering
    \includegraphics[width=\textwidth]{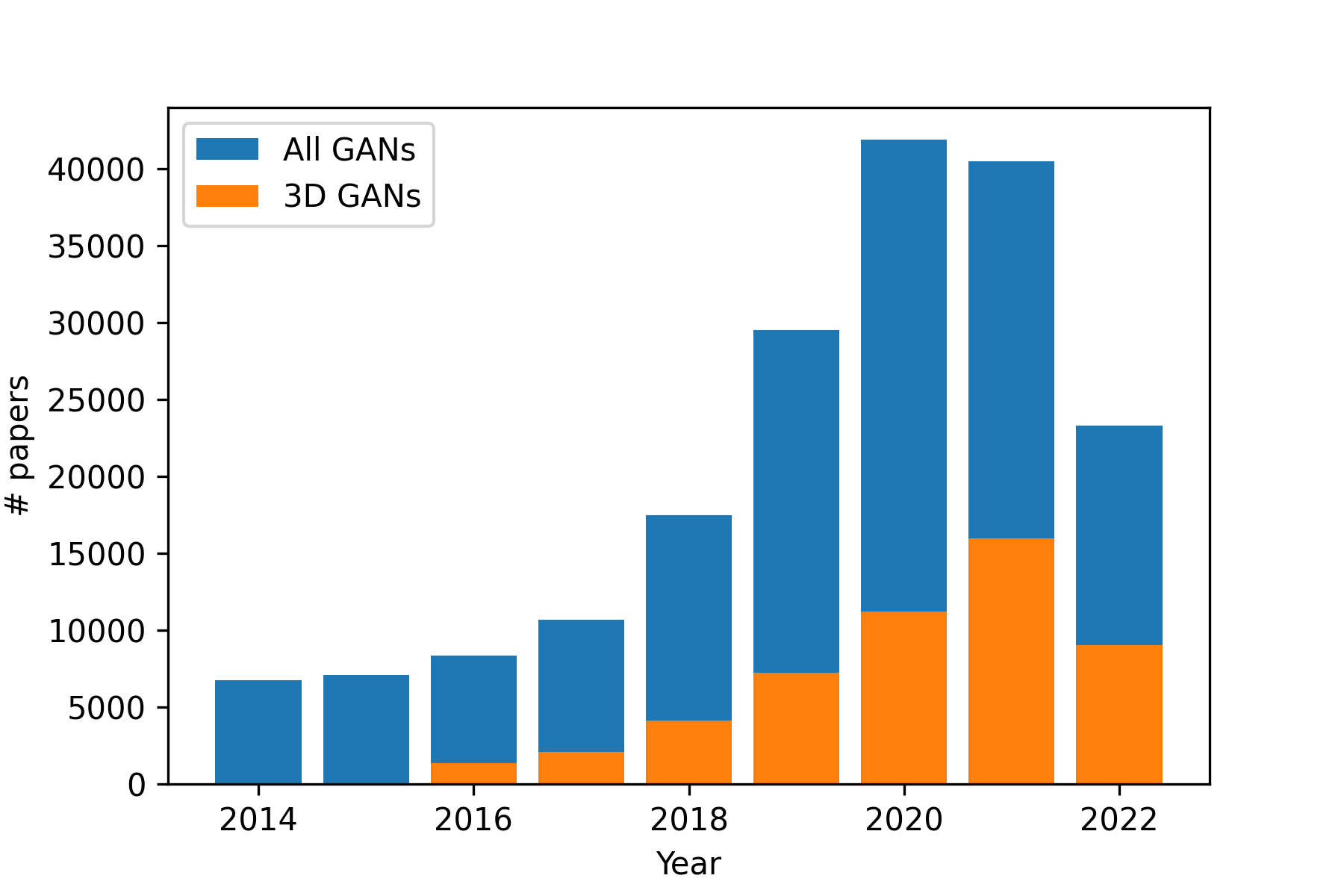}
    \caption{Comparative Year-wise break-up of number of papers published on All GANs and 3D GANs.}
    \label{fig:gan_num}
\end{figure}

In this study, we aim to understand how the research interest for GANs has changed over the years, specifically for 2D and 3D GANs. For this, we look at the number of Google Scholar results obtained within a given year for all GANs and 3D GANs. The results are summarised in tabular form in table \ref{tab:year_gan} and in graphical form in fig \ref{fig:gan_num}. Do note that the data for the year 2022 is incomplete as only the data till June is included in the analysis.

\begin{table}
    \caption{Comparative Year-wise break-up of number of papers published on All GANs and 3D GANs.}
    \label{tab:year_gan}
    \begin{tabular}{p{0.1\textwidth}|p{0.1\textwidth}|p{0.1\textwidth}}
    \toprule
    \textbf{Year} & \textbf{All GANs} & \textbf{3D GANs}\\
    \midrule
    2014 & 6,760  & 0\\
    \hline
    2015 & 7,080  & 0 \\
    \hline
    2016 & 8,340  & 1,380\\
    \hline
    2017 & 10,700 & 2,100\\
    \hline
    2018 & 17,500 & 4,150\\
    \hline
    2019 & 29,500 & 7,220\\
    \hline
    2020 & 41,900 & 11,200\\
    \hline
    2021 & 40,500 & 16,000\\
    \hline
    2022* & 23,300 & 9,020\\
    \bottomrule
    \end{tabular}
\end{table}

\section{Conclusion} \label{la_fin}
Research in Generative Adversarial Networks has been an exciting area for many years and will keep many scientists and engineers busy. In this work, we started with the GAN idea. We explored its latent space with the help of specific latent directions and latent walks to conclude that understanding and local analysis of the latent space geometry and semantics unlocks several opportunities leading to several novel applications. We then introduced 3D GANs, justified why we need them, discussed some improvements to the original 3D-GAN, understood and weighed different 3D representations against one another with the help of domain translation, and classified several 3D GANs based on the task they perform. We then moved the discussion towards finding better ways to train GANs and discussed several shortcomings of both 2D and 3D GANs while indicating future research. We hope this work supplements the interest in GANs. 

\bibliographystyle{acm}
\bibliography{refs}

\end{document}